\documentclass[11pt]{article}
\usepackage{acl2005}
\usepackage{times}

\newcommand{\webtilde}{{\char`\~}}

\newcommand{\problem}{rating-inference problem\xspace}
\newcommand{\problemGeneric}{rating inference\xspace}

\newcommand{\dataitem}{x}
\newcommand{\doc}{\dataitem}

\newcommand{\dataitemalt}{y}

\newcommand{\labelvar}{\ell}

\newcommand{\distvar}{d}
\newcommand{\distfn}{f}
\newcommand{\unif}{\hat{\distfn}}
\newcommand{\plusidentity}{$+$}
\newcommand{\pluspotts}{$\hat{+}$}
\newcommand{\genericnumneighbors}{k}
\newcommand{\neighbors}[1]{nn_{#1}}
 
\newcommand{\weight}{{\it sim}}
\newcommand{\labeldistfn}{d}
\newcommand{\pref}{\pi}
\newcommand{\pospref}[1]{\pref_{+}(#1)}
\newcommand{\labelpref}[2]{\pref(#1,#2)} 

\newcommand{\prefova}[2]{\pref^{{\rm ova}}(#1,#2)}
\newcommand{\prefreg}[2]{\pref^{{\rm reg}}(#1,#2)}
\newcommand{\regfn}{g}

\newcommand{\fnlinkcost}{\sum_{\dataitemalt \in\neighbors{\genericnumneighbors}(\dataitem)}\distfn(\labeldistfn(\labelvar_\dataitem,\labelvar_\dataitemalt)){\weight(\dataitem,\dataitemalt) }}
\newcommand{\ftlinkcost}{\labelpref{\dataitem}{\labelvar_\dataitem}}
\newcommand{\coeff}{\alpha}

\newcommand{\grad}{scale\xspace} 

\newcommand{\gradoc}{\grad dataset\xspace}

\newcommand{\polsent}{sentence polarity dataset\xspace}

\newcommand{\APSbase}{PSP\xspace}
\newcommand{\APSabbrev}{{\APSbase}\xspace} 
\newcommand{\APS}{PSP\xspace}

\newcommand{\APSonly}{0\plusidentity${\rm \APS}^*$\xspace}
\newcommand{\aps}{positive-sentence percentage\xspace}

\newcommand{\apsvecname}[1]{\overrightarrow{{\rm \APSabbrev}(#1)}}
\newcommand{\apsvec}[1]{({\rm \APSabbrev}(#1),  1 -{\rm \APSabbrev}(#1))}

\newcommand{\Binaryclass}{Discretizing binary classification\xspace}

\newcommand{\svmova}{one-vs-all SVM\xspace}
\newcommand{\svmreg}{SVM regression\xspace}

\newcommand{\overlapabbrev}{TO(cos)\xspace}

\newcommand{\three}{three\xspace}
\newcommand{\four}{four\xspace}
\newcommand{\statsig}{significant\xspace}
\newcommand{\statsigly}{{\statsig}ly\xspace}

\newcommand{\ovaabbrev}{ova}
\newcommand{\regabbrev}{reg}
\newcommand{\mpb}[1]{\parbox{.09in}{\begin{center}#1\end{center}}}
\newcommand{\mpba}[1]{\mpb{\tiny{{#1}}}}
\newcommand{\rowb}{$\triangleleft$} 
\newcommand{\rowbaddaps}{\scriptsize{$\blacktriangleleft$}}
\newcommand{\colb}{\tiny{$\triangle$}}
\newcommand{\nodiff}{.}
\newcommand{\nodiffaddaps}{{\bf.}}
\newcommand{\statdiff}[4]{{\renewcommand{\nodiff}{}\renewcommand{\colb}{$\triangleright$}\mbox{#1 #3 #2 [{#4}c]}}}

\newcommand{\raabbrev}{a\xspace}
\newcommand{\rbabbrev}{b\xspace}
\newcommand{\rcabbrev}{c\xspace}
\newcommand{\rdabbrev}{d\xspace}
\newcommand{\Reviewer}{Author\xspace}
\newcommand{\reviewer}{author\xspace}

\newcommand{\rc}{\Reviewer \rcabbrev}

\newcommand{\qanda}[2]{\noindent {\bf Q:~} #1 \\ \noindent {\bf A:~} #2
 \smallskip}

\usepackage{latexsym}
\usepackage{xspace}
\usepackage{graphicx}
\usepackage{amssymb}

\title{Seeing stars: Exploiting class relationships for sentiment categorization with respect to
  rating scales}

\author{$\mbox{Bo Pang}^{1,3}$ \and $\mbox{Lillian Lee}^{1,2,3}$ \\
 (1) Department of Computer Science, Cornell
  University \\
  (2) Language Technologies Institute, Carnegie Mellon University \\
  (3) Computer Science Department, Carnegie Mellon University}
\date{}

\begin{document}

\maketitle

\begin{abstract}
We address 
the {\em \problem}, 
wherein
rather than simply 
decide whether a review is ``thumbs up''
or ``thumbs down'', as in previous sentiment analysis work, 
one must determine an author's evaluation with respect to a 
 multi-point
scale (e.g., one to
 five ``stars'').
This task represents an interesting twist on
standard
multi-class text categorization because
there are several different degrees of similarity between class 
labels; 
for example, 
``three stars'' is intuitively closer to ``four stars'' than
to ``one star''.

We first 
evaluate human performance at the task.
Then,
we 
apply a meta-algorithm, based on a {\em metric labeling} 
formulation of the problem, that 
alters a given $n$-ary classifier's output 
in an explicit attempt to ensure that
similar items receive similar labels.
We show that the meta-algorithm can provide significant improvements
over both 
multi-class and regression versions of SVMs
when we
employ a novel  similarity measure appropriate to the problem.

{\bf Publication info:} {\em Proceedings of the ACL}, 2005.

\end{abstract}

\section{Introduction}
\label{sec:intro}

There has recently been a dramatic surge of interest in 
{\em sentiment
analysis}, as more and more people become aware of the scientific
challenges posed and the scope of new applications enabled by
the processing of subjective language.
(The papers collected by Qu, Shanahan, and Wiebe
\shortcite{Qu+Shanahan+Wiebe:04a} form a representative sample of research in
the area.)
Most prior work on the specific problem of categorizing
expressly opinionated text
has focused on the binary  distinction
of positive vs. negative 
\cite{Turney:02a,Pang+Lee+Vaithyanathan:02a,Dave+Lawrence+Pennock:03a,Yu+Hatzivassiloglou:03a}.
But it is often helpful to have more
information than this binary distinction provides, 
especially if one is ranking items by recommendation  or
comparing several reviewers' 
opinions:  example applications include collaborative filtering and deciding which
conference submissions to accept.

Therefore, 
in this paper we consider generalizing 
to finer-grained 
{\em scales}:
rather than
just determine whether a review is ``thumbs up'' or not, we 
attempt to 
infer 
the author's implied
numerical rating, such as ``three stars'' or ``four stars''.
Note that this differs from identifying opinion {\em strength}
\cite{Wilson+Wiebe+Hwa:04a}: 
rants and raves have the
same strength but represent opposite evaluations, 
and referee forms often allow one to indicate that one is very
confident (high strength) that a conference submission is mediocre
(middling rating).
Also, our task differs from {\em ranking} not only because one can be
given a single item to classify (as opposed to a set
of items to be ordered relative to one another), but because 
there are settings in which classification is harder than ranking, and
vice versa.

One can apply standard  $n$-ary
classifiers or regression to this {\em \problem};
independent work by \newcite{Koppel+Schler:05a} considers such methods.
But an alternative approach that explicitly incorporates information
about item similarities together with label similarity information (for instance, ``one star'' is closer to ``two stars''
than to ``four stars'') is to think of the task as one of {\em metric
  labeling} \cite{Kleinberg+Tardos:02a}, where label relations are
encoded via a distance metric.
This
observation yields a meta-algorithm, applicable to both 
semi-supervised (via graph-theoretic techniques) and supervised
settings,  that  alters a given 
$n$-ary classifier's output so that similar 
items tend to be assigned similar labels.

In what follows, we first demonstrate that humans can 
discern 
relatively
small differences in (hidden) evaluation scores, 
indicating that
\problemGeneric is indeed a meaningful task.  
We then present three types of algorithms --- one-vs-all, regression,
and metric labeling ---  that 
can be distinguished by how explicitly they attempt to leverage
similarity between items and between labels.
Next, we consider  what item similarity
measure to apply, proposing one based on the {\em \aps}.
Incorporating 
this new measure within the
metric-labeling framework is shown to often provide significant improvements over
the other algorithms.

We hope that some of the insights derived here might apply to other
scales for text classifcation that have been considered, such as
clause-level opinion strength
\cite{Wilson+Wiebe+Hwa:04a}; affect types like disgust
\cite{Subasic+Huettner:01a,Liu+Lieberman+Selker:03a}; reading
level \cite{Collins-Thompson+Callan:04a}; and urgency or criticality \cite{Horvitz+Jacobs+Hovel:99a}.

\section{Problem validation and formulation}  
\label{sec:validate}

We first ran a small pilot study on human subjects in order to establish a
rough idea of what a reasonable classification granularity is:
if even people cannot
accurately infer labels with respect to
a five-star scheme with half stars, say, then
we cannot expect a learning algorithm to do so.  Indeed, some potential obstacles to
accurate \problemGeneric include lack of calibration
(e.g., what
an understated \reviewer intends as high praise may seem lukewarm), \reviewer
inconsistency at assigning fine-grained ratings, and ratings not
entirely supported by the text
\footnote{For example, the critic Dennis Schwartz writes that ``sometimes the review itself [indicates] the letter grade should have been higher or lower, as the
  review might fail to take into consideration my overall impression
  of the film --- which I hope to capture in the grade''
(http://www.sover.net/{\webtilde}ozus/cinema.htm).}.

\begin{table}[t]
\begin{tabular}{|l||
r@{\%\hspace*{.04in}}|
r@{\% (}r@{)\hspace*{.04in}}|
r@{\% (}r@{)\hspace*{.04in}}|} \hline
Rating diff.  &  \multicolumn{1}{c|}{Pooled} & 
\multicolumn{2}{c|}{Subject 1} & \multicolumn{2}{c|}{Subject 2} \\ \hline
{$3$ or more}     &   100  &   100   &   35   &   100  &   15  \\
2 (e.g., 1~star)   &   83  &   77   &   30   &   100  &   11  \\
1  (e.g., $\frac{1}{2}$~star)   &   69  &   65   &   57   &   90  &   10  \\
 {0}   &   55   &   47   &   15   &   80  &   5  \\ \hline
\end{tabular}

 \caption{\label{tab:validation}
Human accuracy at determining relative positivity. Rating differences
are given in  ``notches''.  Parentheses
enclose the number of pairs attempted. 
 }
\end{table}

For data, we first collected Internet movie reviews in English from four
authors, removing explicit rating indicators from each
document's text automatically.
Now, while the obvious experiment would be to ask subjects to guess the rating
that a review represents, doing so would force us to specify
a fixed rating-scale granularity in advance.  Instead, we examined
people's ability to discern {\em relative differences}, because
by varying the rating differences represented by the test instances,
we can evaluate multiple granularities in a single experiment.
Specifically, at intervals over a number of weeks, 
we authors (a non-native and a native speaker of English)
examined
pairs of reviews,
attemping to determine whether the first review in each pair was (1) {more positive than}, (2) {less
positive than}, or (3) {as positive as} 
the second.  
The texts in any particular
review pair were taken from the same author to factor out the effects
of 
cross-author divergence.

As Table \ref{tab:validation} shows,
 both subjects performed
perfectly when the rating separation was at least 3 ``notches''  in the
original scale (we define a notch as a half star in a four- or five-star scheme
 and 10 points in a 100-point scheme). Interestingly, although human performance drops as rating
difference decreases, even at a one-notch  separation, both subjects  handily outperformed the random-choice baseline
of 33\%. However, there was large variation in accuracy between subjects.
\footnote{
One contributing factor 
may be that the subjects viewed disjoint document
sets,  since we wanted to 
maximize experimental coverage of the types of document
pairs within each difference class.
We thus cannot report inter-annotator agreement, but since
our goal is to recover a reviewer's ``true''
recommendation,  reader-author
agreement is more relevant.

While another factor might be degree of English  fluency,
in an informal experiment (six subjects viewing the same
three pairs), native English speakers made the only two errors.
}

Because of this variation, we defined two different
classification regimes.
From the evidence above, a {\bf three-class} task 
(categories 0, 1, and
2 --- essentially ``negative'', ``middling'', and
``positive'', respectively) seems like one that most people would
do quite well at  (but we should not assume
100\% human accuracy: according to our one-notch results, people
may misclassify borderline cases like 2.5 stars).
Our study also suggests that  people could do at least fairly well at
distinguishing full stars in a zero- to four-star scheme.  However, when we
began to  construct five-category
datasets for each of our four authors (see below),  we found that in
each case,  either the most negative or the most positive class (but
not both)
contained only about 5\% of the documents.  To make the classes more
balanced, we folded these minority
classes into the adjacent class, thus arriving at a {\bf
four-class} problem (categories 0-3, increasing in positivity).
Note that the four-class problem seems to offer more
possibilities for leveraging class relationship information than the
three-class setting, since
it involves more class pairs.
Also,  even the two-category version of the 
\problem for movie reviews has proven quite challenging for many
automated classification techniques 
\cite{Pang+Lee+Vaithyanathan:02a,Turney:02a}.

We applied the above two labeling schemes to  a {\bf
  \gradoc}\footnote{Available at
http://www.cs.cornell.edu/People/pabo/movie-review-data as \gradoc
v1.0.} containing four corpora of movie reviews.  
All reviews were
automatically  pre-processed to remove both explicit rating indicators and
objective sentences; the motivation for the latter step is that it has
previously aided 
positive vs. negative classification \cite{Pang+Lee:04a}.
All of the 1770, 902, 1307, or 1027 documents in a given corpus were
written by the same author.
This 
decision facilitates interpretation of the results,
since it factors out the effects of different choices of
methods for calibrating  authors' scales.
\footnote{ 
  From the Rotten Tomatoes website's FAQ:
  ``star systems are not consistent between critics.  For critics like
  Roger Ebert and James Berardinelli, 2.5 stars or lower out of 4
  stars is always negative. For other critics, 2.5 stars can either be
  positive or negative. Even though Eric Lurio uses a 5 star system,
  his grading is very relaxed. So, 2 stars can be positive.''  
Thus,
 calibration may sometimes require strong familiarity
  with the authors involved, as anyone who has ever needed
  to reconcile conflicting referee reports probably knows. }
We point out that it is possible to gather \reviewer-specific information
in some practical applications: for instance, systems that 
use selected authors (e.g.,  the Rotten Tomatoes
movie-review website ---  where, we note, not all authors
provide explicit ratings) could require that someone
submit  rating-labeled samples
of newly-admitted authors' work.  Moreover, our
results  at least partially generalize to mixed-author situations
(see Section \ref{sec:disc}).

\section{Algorithms}
\label{sec:method}

Recall that the problem we are considering is  multi-category
classification in which the labels can be naturally mapped to 
a metric space 
(e.g., points on a line);
for simplicity, we assume 
the distance metric
$\labeldistfn(\labelvar,\labelvar') =
|\labelvar-\labelvar'|$ throughout.
In this section, we present three approaches to this problem in order of increasingly explicit use of pairwise similarity
information between items and between labels.  In order to make
comparisons between these methods meaningful, we base all three of
them on Support Vector Machines (SVMs)
 as implemented in Joachims'
\shortcite{Joachims:99a} ${\rm SVM}^{light}$ package.

\subsection{One-vs-all}
\label{sec:naryclass}

The standard SVM formulation applies only to binary classification.
{\em One-vs-all} (OVA) \cite{Rifkin+Klautau:04a} is 
a common extension to the $n$-ary case.
Training consists of building, for each label $\labelvar$, an SVM
binary classifier distinguishing label $\labelvar$ from
``not-$\labelvar$''.
We consider the final output to be a label preference function
$\prefova{\dataitem}{\labelvar}$, defined as the 
signed distance of (test) item $\dataitem$ to the $\labelvar$ side of the
$\labelvar$ vs. not-$\labelvar$ decision plane.

Clearly, OVA makes no explicit use of pairwise label or item
relationships.  However, it can perform well if each class
exhibits sufficiently distinct language; see Section \ref{sec:mix} for
more discussion.

\subsection{Regression}
\label{method:svm-reg}

Alternatively, we can take a {\em regression} perspective by
assuming that the labels come from a discretization of a
continuous function $\regfn$
mapping from the feature space to a metric space.
\footnote{We discuss the {\em ordinal} regression variant in Section \ref{sec:relwork}.}
If we choose $\regfn$ from a family of sufficiently ``gradual''
functions, then similar items necessarily receive similar labels.
In particular, we consider {\em
linear, $\varepsilon$-insensitive} SVM regression
\cite{Vapnik:95a,Smola+Schoelkopf:98a}; the  idea is to find the
hyperplane that best fits the training data, 
but where training points whose labels are within distance $\varepsilon$ of the
hyperplane incur no loss.
Then, for (test) instance $\dataitem$, the label
preference function $\prefreg{\dataitem}{\labelvar}$ is 
the negative
of the
distance  between $\labelvar$ and the 
value  predicted for
$\dataitem$ by the fitted hyperplane function.

\newcite{Wilson+Wiebe+Hwa:04a} used SVM regression to
classify clause-level strength of opinion, reporting that it provided lower
accuracy
than other methods.
However, independently of our work, \newcite{Koppel+Schler:05a} found
that applying linear regression to classify documents (in a different
corpus than ours) with respect to
a three-point rating scale provided greater accuracy than OVA SVMs and
other algorithms.

\subsection{Metric labeling}
\label{sec:aps}

Regression {\em implicitly} encodes the ``similar items, similar labels''
heuristic, in that one can restrict consideration to ``gradual'' functions.
But we can
also think of our task as a {\em metric labeling} problem
\cite{Kleinberg+Tardos:02a}, 
a special case of the maximum {\em a posteriori}
estimation problem for Markov random
fields,
to {\em explicitly} encode our desideratum.
Suppose we have an initial label preference function
$\labelpref{\dataitem}{\labelvar}$, perhaps computed via one of the two
methods described above.  Also,
 let $\labeldistfn$ be a distance
metric on labels, and let
$\neighbors{\genericnumneighbors}(\dataitem)$ denote the $\genericnumneighbors$ nearest neighbors
of item $x$
according to some item-similarity function  $\weight$. 
Then, it is quite natural to pose our problem as finding a mapping of instances
$\dataitem$ to labels  $\labelvar_\dataitem$ (respecting the
original labels of the training instances)  that minimizes
\hspace*{-.15in}\begin{minipage}{1.5in}
\[
\sum_{\dataitem \in \mbox{test}} 
\left[- \ftlinkcost +
  \coeff
\fnlinkcost
\right],
\label{eq:cost}
\]
\end{minipage} \\
where $\distfn$ is monotonically increasing (we chose
$\distfn(\distvar)=\distvar$ unless otherwise specified
)
and $\coeff$ is a 
trade-off and/or scaling parameter.
(The inner summation is familiar from work in {\em locally-weighted
learning}
\footnote{
If we ignore the $\pref(\dataitem,\labelvar)$ term, 
different choices of  $\distfn$ 
correspond to different versions of nearest-neighbor learning, e.g.,
majority-vote, weighted average of labels, or weighted median of labels.
}  \cite{Atkeson+Moore+Schaal:97a}.)
In a sense, we are using explicit item and label similarity
information to increasingly penalize
the initial classifier 
as it assigns more divergent labels to similar items.

In this paper, we 
only report 
supervised-learning experiments in which the nearest neighbors for
any given test item were drawn from the training set alone.  In such a
setting, the labeling decisions for different test items are
independent, so that solving the requisite optimization problem is simple.

\paragraph{Aside: transduction} The above formulation also allows for {\em transductive}
semi-supervised learning as well, in that we could allow nearest
neighbors to come from both the training and test sets.  We intend to
address this case in future work, since 
there are important settings in which one has a
small 
number of labeled reviews and a large number of unlabeled reviews,
in which case considering similarities between unlabeled texts could
prove quite helpful.
In full generality,
the corresponding multi-label optimization problem 
is intractable,
but 
for many  families of $\distfn$ functions (e.g., convex) there exist practical exact or
approximation algorithms based
on techniques for finding
{\em minimum s-t cuts} in graphs 
\cite{Ishikawa+Geiger:98a,Boykov+Veksler+Zabih:99a,Ishikawa:03a}.
Interestingly, previous sentiment analysis research found that 
a minimum-cut
formulation 
for the binary subjective/objective distinction
yielded good results \cite{Pang+Lee:04a}.  Of course, there are many
other related semi-supervised learning algorithms that we would like
to try as well; see  \newcite{Zhu:05a} for a  survey.

\section{Class struggle: finding a label-correlated item-similarity function}
\label{sec:mix}

We 
need  to specify an item similarity function $\weight$ to
use the metric-labeling formulation described in Section \ref{sec:aps}.  We could, as is commonly
done, 
employ 
a term-overlap-based measure such as the cosine between term-frequency-based document vectors
(henceforth ``\overlapabbrev'').  However, 
Table \ref{tab:vocab-overlap} shows that in aggregate, the
vocabularies of distant classes overlap to a degree surprisingly similar
to that of the vocabularies of nearby classes.
Thus, item similarity
as measured by \overlapabbrev may not correlate well with similarity of
the item's true labels.

\begin{table}[t]
\centering
\begin{tabular}{|l|ccc|}\hline
 & \multicolumn{3}{c|}{Label difference:} \\
 & \multicolumn{1}{c}{1} & \multicolumn{1}{c}{2} &
 \multicolumn{1}{c|}{3} \\ \hline\hline
Three-class data & 37\% & 33\% & --- \\ \hline
Four-class data & 34\% & 31\% & 30\%\\\hline
\end{tabular}
\caption{\label{tab:vocab-overlap} Average over authors and class
 pairs of between-class vocabulary
 overlap 
 as the class labels of the pair grow farther apart. }
\end{table}

We can potentially develop a more useful similarity metric by
asking ourselves what, intuitively, accounts for the label relationships that we seek
to exploit.
A simple hypothesis
is that ratings can be determined by the {\em \aps
(\APSabbrev)} of a text, 
i.e., the number of positive sentences divided by the number of subjective sentences.
(Term-based versions of this premise have 
motivated much sentiment-analysis work for 
over a decade
\cite{Das+Chen:01a,Tong:01a,Turney:02a}.)
But counterexamples are easy to construct:  reviews can contain
off-topic opinions, or recount many positive aspects before
describing a fatal flaw.  

We therefore tested the hypothesis as follows.
To avoid the need to hand-label sentences as positive or negative, we
first created 
a {\em \polsent}
\footnote{Available at
 http://www.cs.cornell.edu/People/pabo/movie-review-data as \polsent
 v1.0.}
consisting of 
10,662  movie-review
``snippets'' (a striking extract usually one sentence long) downloaded from
www.rottentomatoes.com; 
each snippet was
labeled with 
its source review's 
label (positive or negative)
as provided by Rotten Tomatoes.
Then, we trained a Naive Bayes classifier on this data set and applied
it to our \gradoc to identify the positive sentences
(recall that objective sentences were already removed).

\begin{figure}[th]
\includegraphics[width=3.3in]{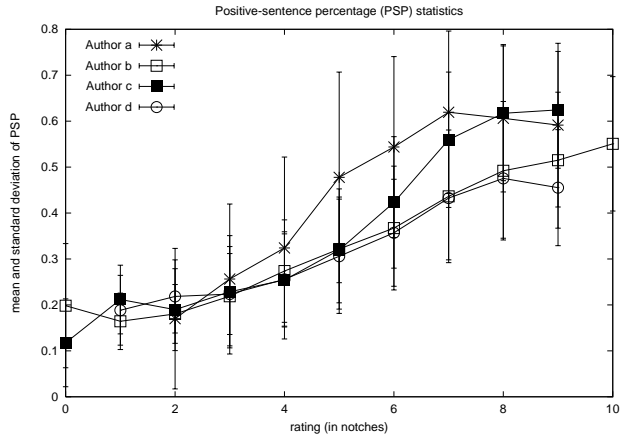}
\caption{\label{fig:APS}
Average and standard deviation of \APSabbrev for reviews expressing different ratings. 
}
\end{figure}

Figure \ref{fig:APS}
shows
that all four authors tend to exhibit a higher \APSabbrev when they
write a more positive review,
and we expect that most typical reviewers would follow suit. Hence, 
\APSabbrev appears to be a promising basis for computing document
similarity for our rating-inference task.
In particular, 
we defined
$\apsvecname{\dataitem}$ to be the 
two-dimensional 
vector $\apsvec{\dataitem}$,  and then set
the item-similarity function
required by the metric-labeling optimization function (Section \ref{sec:aps}) to
$\weight(\dataitem,\dataitemalt) =
\cos\left(\apsvecname{\dataitem},\apsvecname{\dataitemalt}\right).$
\footnote{While admittedly we initially chose this function because it was
convenient to work with cosines, {\em post hoc} analysis revealed that
the corresponding metric space ``stretched'' certain distances in a
useful way.}

But before proceeding, we note that it is possible that similarity
information might  yield no extra benefit at all.  For
instance, we don't need it if
we can reliably identify each class just from some set of
distinguishing terms. 
If we define such terms as frequent ones
($n\geq 20$)  that appear in a single class 50\% or more of the time,
then we do find many instances;
some examples for one \reviewer are:
``meaningless'', ``disgusting'' (class 0); ``pleasant'',
``uneven'' (class 1); and ``oscar'', ``gem'' (class 2) for the
\three-class case, and, in the \four-class case, ``flat'', ``tedious''
(class 1) versus ``straightforward'', ``likeable'' (class 2).  Some
unexpected distinguishing terms for this \reviewer are ``lion'' for
class 2 (\three-class case), and for class 2 in the \four-class case, ``jennifer'', for a wide
variety of Jennifers.

\section{Evaluation}
\label{sec:eval}

\newcommand{\tabsvmova}{SVM one-vs-all (ova)}
\newcommand{\tabsvmposratio}{ova\plusidentity\APS}
\newcommand{\tabsvmregdisc}{SVM regression (reg)}
\newcommand{\tabngsvmregressionposratio}{reg\plusidentity\APS}

This section compares the accuracies of the  approaches outlined in
Section \ref{sec:method} on the four corpora comprising  our
\gradoc. (Results using $L_1$ error were qualitatively similar.)
Throughout, when we refer to something as ``significant'', we mean
statistically so
with respect to the paired $t$-test, $p<.05$.

The results that follow
are based on ${\rm SVM}^{light}$'s default parameter settings for SVM regression
and OVA.
Preliminary analysis of the effect of varying the regression parameter
$\varepsilon$ in the \four-class case revealed that the default value
was often optimal.

The notation ``A{\plusidentity}B''  denotes 
metric labeling 
where method A provides the initial
label preference function $\pref$ and B
serves as similarity measure.  
To train, we first select the meta-parameters $\genericnumneighbors$
and $\coeff$ by running 9-fold cross-validation within the training
set.  
Fixing $\genericnumneighbors$
and $\coeff$ 
to those values yielding the best performance, we then re-train A 
(but with SVM parameters fixed, as described above) on the whole
training set. At test time, 
the nearest neighbors of each 
item are also taken from the full
training set.

 \subsection{Main comparison}
\label{sec:acc}

\begin{figure*}[ht]

\hspace*{-.5in}
\begin{tabular}{cc}
{\bf Average accuracies, three-class data} & {\bf Average accuracies,
  four-class data} \\ 
\includegraphics[width=3.5in]{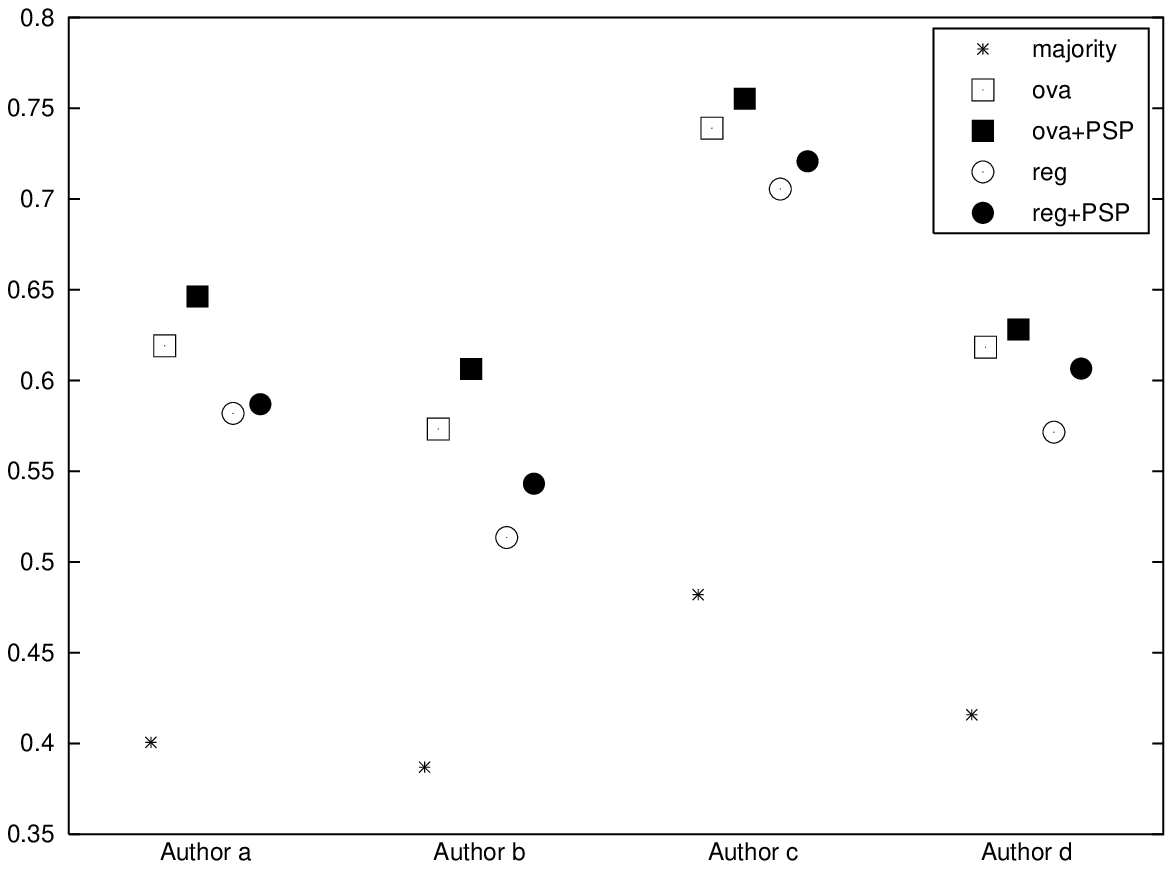} &
\includegraphics[width=3.5in]{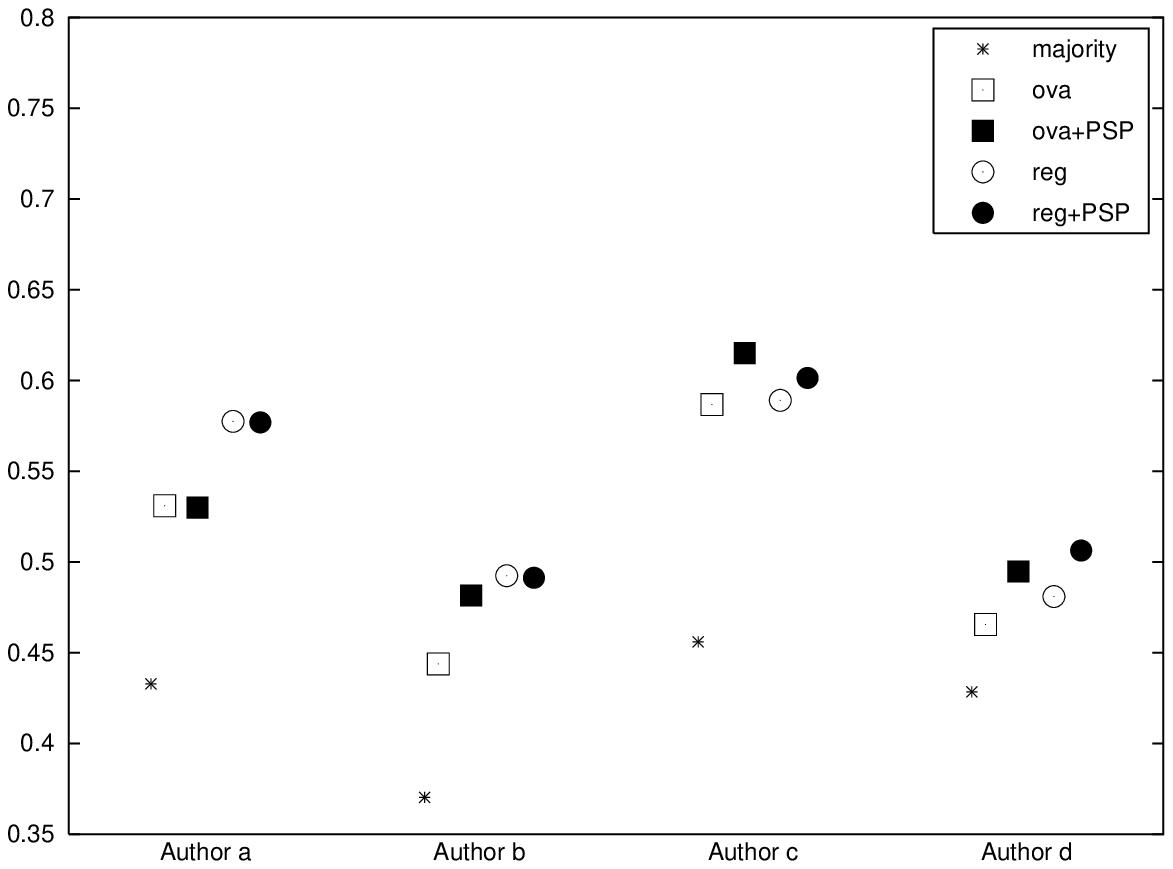} \\
\end{tabular}

Average ten-fold cross-validation accuracies. Open icons:
SVMs in either one-versus-all (square)
or regression (circle) mode; dark versions:
metric labeling using the corresponding SVM  together with the \aps
(\APS).
The $y$-axes of the two plots are aligned. 

\vspace*{.35in}

\hspace*{-.3in}
\begin{tabular}{cp{.1in}c}
{\bf Significant differences, three-class data}  
& 
&
{\bf Significant differences, four-class data}\\
\fbox{\begin{tabular}[b]{l|*{4}{c}}
 & \ovaabbrev &\ovaabbrev+\APS &\regabbrev
 &\regabbrev+\APS \\ 
  & \mpba{\raabbrev}\mpba{\rbabbrev}\mpba{\rcabbrev}\mpba{\rdabbrev}
&  \mpba{\raabbrev}\mpba{\rbabbrev}\mpba{\rcabbrev}\mpba{\rdabbrev}
& \mpba{\raabbrev}\mpba{\rbabbrev}\mpba{\rcabbrev}\mpba{\rdabbrev}
& \mpba{\raabbrev}\mpba{\rbabbrev}\mpba{\rcabbrev}\mpba{\rdabbrev} \\ \hline

\ovaabbrev      &       &\mpb{\colb}\mpb{\colb}\mpb{\colb}\mpb{\nodiff}       &\mpb{\rowb}\mpb{\rowb}\mpb{\rowb}\mpb{\rowb}   &\mpb{\nodiff}\mpb{\rowb}\mpb{\nodiff}\mpb{\nodiff}      \\
\ovaabbrev+\APS  &\mpb{\rowbaddaps}\mpb{\rowbaddaps}\mpb{\rowbaddaps}\mpb{\nodiff}       &       &\mpb{\rowb}\mpb{\rowb}\mpb{\rowb}\mpb{\rowb}   &\mpb{\rowb}\mpb{\rowb}\mpb{\rowb}\mpb{\nodiff}      \\
\regabbrev      &\mpb{\colb}\mpb{\colb}\mpb{\colb}\mpb{\colb}   &\mpb{\colb}\mpb{\colb}\mpb{\colb}\mpb{\colb}   &       &\mpb{\nodiff}\mpb{\colb}\mpb{\nodiff}\mpb{\colb}  \\
\regabbrev+\APS  &\mpb{\nodiff}\mpb{\colb}\mpb{\nodiff}\mpb{\nodiff}       &\mpb{\colb}\mpb{\colb}\mpb{\colb}\mpb{\nodiff}       &\mpb{\nodiff}\mpb{\rowbaddaps}\mpb{\nodiff}\mpb{\rowbaddaps} &        \\

\end{tabular}
}
& 
&
\fbox{
\begin{tabular}[b]{l|*{4}{c}}
  & \ovaabbrev &\ovaabbrev+\APS &\regabbrev &\regabbrev+\APS \\ 
  & \mpba{\raabbrev}\mpba{\rbabbrev}\mpba{\rcabbrev}\mpba{\rdabbrev}
&  \mpba{\raabbrev}\mpba{\rbabbrev}\mpba{\rcabbrev}\mpba{\rdabbrev}
& \mpba{\raabbrev}\mpba{\rbabbrev}\mpba{\rcabbrev}\mpba{\rdabbrev}
& \mpba{\raabbrev}\mpba{\rbabbrev}\mpba{\rcabbrev}\mpba{\rdabbrev} \\ \hline

\ovaabbrev      &       &\mpb{\nodiff}\mpb{\colb}\mpb{\colb}\mpb{\colb}       &\mpb{\colb}\mpb{\colb}\mpb{\nodiff}\mpb{\nodiff}   &\mpb{\colb}\mpb{\nodiff}\mpb{\nodiff}\mpb{\colb}  \\
\ovaabbrev+\APS  &\mpb{\nodiff}\mpb{\rowbaddaps}\mpb{\rowbaddaps}\mpb{\rowbaddaps}       &       &\mpb{\colb}\mpb{\nodiff}\mpb{\nodiff}\mpb{\nodiff}       &\mpb{\colb}\mpb{\nodiff}\mpb{\nodiff}\mpb{\nodiff}      \\
\regabbrev      &\mpb{\rowb}\mpb{\rowb}\mpb{\nodiff}\mpb{\nodiff}   &\mpb{\rowb}\mpb{\nodiff}\mpb{\nodiff}\mpb{\nodiff}       &       &\mpb{\nodiff}\mpb{\nodiff}\mpb{\nodiff}\mpb{\nodiff}   \\
\regabbrev+\APS  &\mpb{\rowb}\mpb{\nodiff}\mpb{\nodiff}\mpb{\rowb}   &\mpb{\rowb}\mpb{\nodiff}\mpb{\nodiff}\mpb{\nodiff}       &\mpb{\nodiffaddaps}\mpb{\nodiffaddaps}\mpb{\nodiffaddaps}\mpb{\nodiffaddaps}   &       \\

\end{tabular}
}\\ 
\end{tabular}

\vspace*{.1in}

Triangles point towards \statsigly better algorithms for the results
plotted above.
Specifically, if the difference between a row and a column algorithm for a given author dataset (\raabbrev, \rbabbrev,
\rcabbrev, or \rdabbrev) is \statsig, a triangle points to
the better one; otherwise, a dot (.) is shown.
Dark icons highlight the effect of adding \APS information via metric labeling.

\caption{\label{fig:acc}Results for main experimental comparisons.}
\end{figure*}

Figure \ref{fig:acc} summarizes our average 10-fold cross-validation
accuracy
results.
We first observe from the plots that all the algorithms described in Section
\ref{sec:method} 
always 
definitively outperform
the simple baseline of  predicting the
majority class, although the improvements are smaller in the
\four-class case.
Incidentally, the data was distributed in  such a way that the absolute performance of the  baseline
itself does not change much between the \three- and \four-class case
(which implies that the \three-class datasets were relatively more balanced);
and  \rc's 
datasets seem noticeably easier than the others.

We now examine the effect of implicitly using label and item similarity.
In the \four-class case, regression
performed better than OVA (\statsigly so for
two {\reviewer}s, as shown in
the righthand table);
but for the \three-category task, OVA 
\statsigly outperforms 
regression
for all four
authors.
One might initially interprete this ``flip'' as showing that in the \four-class
scenario, item and label similarities provide a 
richer source of information
relative to class-specific characteristics, especially since for the
non-majority classes there is less data available; whereas in the \three-class
setting the categories are better modeled as quite distinct entities.

However, the \three-class results for metric labeling on top of OVA and
regression (shown in Figure \ref{fig:acc} by black versions of the
corresponding icons) show that employing explicit similarities always
improves results, often to a \statsig degree, and 
yields the best overall accuracies.  Thus,
we 
{\em can} in fact effectively exploit similarities in the \three-class
case.
Additionally, in both the \three- and \four- class scenarios,
metric labeling often brings the performance of the weaker base method
up to that of the stronger one (as indicated by the
``disappearance'' of upward triangles in corresponding table rows), and 
never hurts performance \statsigly.

In the \four-class case, metric labeling and regression seem roughly
equivalent.  One possible interpretation is that the relevant
structure of the problem is already 
captured by linear
regression (and perhaps a different kernel for regression would have
improved its \three-class performance).  However, according to
additional experiments we ran
in the \four-class situation, 
the test-set-optimal parameter settings for
metric labeling would have produced \statsig improvements,
indicating there may be greater potential for our
framework.  At any rate, we view the fact that 
metric labeling
performed quite well for both rating scales 
as a 
definitely positive result.

\subsection{Further discussion}
\label{sec:disc}

\qanda{Metric labeling looks like it's
  just combining SVMs with nearest neighbors, and classifier
  combination often improves performance.  Couldn't we get the same
  kind of results by  combining SVMs with any other
  reasonable method?}  { No.  For example, if we take the strongest
  base SVM method for initial label preferences, but replace
  \APSabbrev with the term-overlap-based cosine (\overlapabbrev),
  performance often drops \statsigly.  This
  result, which is in accordance with Section \ref{sec:mix}'s data, suggests
  that choosing an item similarity function that correlates well with label
  similarity is important.
(\statdiff{\ovaabbrev\plusidentity\APS}{\ovaabbrev\plusidentity\overlapabbrev}{\rowb\rowb\rowb\rowb}{3}; 
\statdiff{\regabbrev\plusidentity\APS}{\regabbrev\plusidentity\overlapabbrev}{\nodiff\nodiff\nodiff\rowb}{4})
}

\qanda{Could you explain that  notation, please?}
{
  Triangles point toward the \statsigly better algorithm for some
  dataset.   For instance, ``\statdiff{M}{N}{\rowb\rowb\colb\nodiff}{3}'' means,
``In the 3-class task, method M is \statsigly better than N for two
  author datasets and \statsigly worse
for one dataset (so the algorithms were statistically
  indistinguishable on the remaining dataset)''. 
When the algorithms being compared are statistically indistinguishable
on all four datasets (the ``no triangles'' case), we indicate this
with an equals sign (``='').
}

\qanda{Thanks.  Doesn't Figure \ref{fig:APS} show that the \aps would
  be a good classifier even in isolation, so metric labeling isn't necessary?}
{
No.
Predicting class labels directly from the \APS value via trained
  thresholds 
isn't as effective
(\statdiff{\ovaabbrev\plusidentity\APS}{threshold \APS}{\rowb\rowb\rowb\rowb}{3};
\statdiff{\regabbrev\plusidentity\APS}{threshold \APS}{\rowb\nodiff\rowb\nodiff}{4}).

Alternatively, we could use only the \APSabbrev component of metric labeling by
setting the label preference function to the constant function 0,
but even with {\em test-set-optimal} parameter settings, doing so
underperforms the {\em trained} metric labeling algorithm with access
to an initial SVM classifier
(\statdiff{\ovaabbrev\plusidentity\APS}{\APSonly}{\rowb\rowb\rowb\rowb}{3};
\statdiff{\regabbrev\plusidentity\APS}{\APSonly}{\rowb\nodiff\rowb\nodiff}{4}).
}

\qanda{What about using \APSabbrev as one of the features for input to a standard
  classifier?}{Our focus is on investigating the
  utility of similarity information.  In our particular
  rating-inference setting, it  so happens that the basis for our
  pairwise similarity measure can be
  incorporated as an item-specific feature, but we view this as a
  tangential issue.  That being said, preliminary experiments show
  that metric labeling can be  better, barely (for test-set-optimal
  parameter settings for both algorithms: 
 \statsigly better
  results for one author, \four-class 
case; statistically
  indistinguishable otherwise),
although one needs to determine an appropriate weight for the \APSabbrev
feature 
to get good performance.
}

\qanda{You 
defined the ``metric transformation'' function $\distfn$ as the
identity function $\distfn(\distvar)=\distvar$,
imposing greater loss as the distance between labels assigned to two similar items increases.
Can you do just as well if you penalize all non-equal label
assignments by the same amount, or does the distance between labels really matter?
}
{
You're asking for a comparison to the {\em Potts model}, which 
sets $\distfn$ to the function $\unif(\distvar)=1$ if  $\distvar > 0$, $0$ otherwise.  
In the one setting in which there is a \statsig difference between
the two, the Potts model does worse
(\statdiff{\ovaabbrev\plusidentity\APS}{\ovaabbrev\pluspotts\APS}{\rowb\nodiff\nodiff\nodiff}{3}).
Also, employing 
the 
Potts model generally leads to fewer \statsig
improvements over a chosen base method (compare Figure \ref{fig:acc}'s tables with: \statdiff{\regabbrev\pluspotts\APS}{\regabbrev}{\nodiff\nodiff\nodiff\rowb}{3};
\statdiff{\ovaabbrev\pluspotts\APS}{\ovaabbrev}{\rowb\rowb\nodiff\nodiff}{3};
\statdiff{\ovaabbrev\pluspotts\APS}{\ovaabbrev}{\nodiff\nodiff\nodiff\nodiff$=$}{4};
but note that
\statdiff{\regabbrev\pluspotts\APS}{\regabbrev}{\nodiff\nodiff\nodiff\rowb}{4}).
We note that optimizing the
Potts model in the multi-label case is NP-hard, whereas the 
optimal metric labeling with the identity metric-transformation
function can be efficiently 
obtained (see Section \ref{sec:aps}).
}

\qanda{Your datasets had many labeled reviews and only one
  \reviewer each.  Is your work relevant to settings with many
  {\reviewer}s but very little data for each?
}{As discussed in Section \ref{sec:validate}, it can be quite
  difficult to properly calibrate different {\reviewer}s' scales, since
  the same number of ``stars'' even within what is ostensibly the same
  rating system can mean different things for different {\reviewer}s.
But since you ask: we temporarily turned a blind eye to this serious issue,
creating a collection of 5394 reviews by 496 {\reviewer}s with at most
  80 reviews per \reviewer, where we pretended that our rating
  conversions mapped correctly into a universal rating scheme.
Preliminary results on this 
dataset were actually
comparable to the results reported above, although since we are not
confident in the class labels themselves, more work is needed to
derive a clear analysis of this setting.
(Abusing notation, since we're already playing fast and loose: [3c]:
baseline 52.4\%,  {\regabbrev} 61.4\%, {\regabbrev\plusidentity\APS}
61.5\%, {\ovaabbrev} (65.4\%) $\triangleright$
{\ovaabbrev\plusidentity\APS} (66.3\%);
[4c]: baseline 38.8\%, {\regabbrev} (51.9\%)  $\triangleright$
\regabbrev\plusidentity\APS (52.7\%), {\ovaabbrev} (53.8\%) $\triangleright$ {\ovaabbrev\plusidentity\APS} (54.6\%))

In future work, it would be interesting to determine \reviewer-independent
characteristics that can be used on (or suitably adapted to) data for specific
{\reviewer}s. 
}

\qanda{How about trying ---}{---Yes, there are many 
  alternatives. A few that we tested are described in the Appendix, and
  we  propose some others in the next section.  We should mention that
   we have not yet experimented with {\em all-vs.-all} (AVA), another standard
  binary-to-multi-category classifier conversion method, because we
  wished to
  focus on the effect of omitting pairwise information. In
  independent work on 3-category \problemGeneric for a different corpus, \newcite{Koppel+Schler:05a}  found that
  regression outperformed  AVA, and \newcite{Rifkin+Klautau:04a} argue
  that in principle OVA should do just as well as AVA. But we plan to
  try it out.
 
}

\section{Related work and future directions}
\label{sec:relwork}

In this paper, we addressed the \problem, 
showing the utility of employing label similarity and (appropriate choice of) item
similarity  ---  either implicitly,
through regression, or explicitly and often more effectively, through
metric labeling.

In the future, we would like to apply our methods to other
scale-based classification problems, and explore alternative methods.
Clearly, varying the kernel in SVM regression might yield better results.
Another choice is {\em ordinal regression}
\cite{McCullagh:80a,Herbrich+Graepel+Obermayer:00a}, which only considers the
ordering on labels, rather than any explicit distances between them;
this approach could work well if a good
metric on labels is lacking.  Also, one could use mixture models (e.g., combine 
``positive'' and ``negative'' language models)
to capture class relationships
\cite{McCallum:99a,Schapire+Singer:00a,Takamura+Matsumoto+Yamada:04a}.

We are also interested in framing
multi-class but {\em non}-scale-based categorization problems as metric
labeling tasks.  For example, 
positive vs. negative vs. neutral sentiment
distinctions are sometimes considered in which neutral means
either objective \cite{Engstroem:04a} or a conflation of 
objective with a rating of mediocre
\cite{Das+Chen:01a}. (Koppel and Schler \shortcite{Koppel+Schler:05a}
in independent work also discuss various types of neutrality.)  In either case,
we could apply a metric in which positive and negative are closer to
objective (or objective+mediocre) than to each other. As another
example, hierarchical label relationships can be easily encoded in a
label metric.

Finally, as mentioned in Section \ref{sec:aps}, we would like to
address the transductive setting, in which one has a small amount of
labeled data and uses relationships between unlabeled items, since it is
particularly well-suited to the metric-labeling approach and may be
quite important in practice.

{
\paragraph*{Acknowledgments}  We thank Paul Bennett, Dave Blei, Claire Cardie, Shimon Edelman, Thorsten Joachims,
Jon Kleinberg, Oren Kurland, John Lafferty, Guy Lebanon, Pradeep
Ravikumar, Jerry Zhu, and the anonymous reviewers for many very useful
comments and discussion.  We learned of Moshe Koppel and Jonathan
Schler's work while preparing the camera-ready version of this paper;
we thank them for so quickly answering our request for a pre-print.
Our descriptions of their work are based on that pre-print; we
apologize in advance for any inaccuracies in our descriptions that
result from changes between their pre-print and their final version.
We also thank CMU for its hospitality during the year.  This paper is
based upon work supported in part by the National Science Foundation
(NSF) under grant no.  IIS-0329064 and CCR-0122581;
SRI International under subcontract
no. 03-000211 on their project funded by the Department of the
Interior's National Business Center; and by an Alfred P. Sloan Research
Fellowship. Any opinions, findings, and conclusions or recommendations
expressed are those of the authors and do not necessarily reflect the
views or official policies, either expressed or implied, of any
sponsoring institutions, the U.S. government, or any other entity.  
}

\newcommand{\bibsnip}{\vspace*{-.08in}}

\appendix
\section{Appendix: other variations attempted}
\newcommand{\tabngsvmregknn}{reg+word overlap}
\newcommand{\tabngsvmknn}{ova+word overlap}
\newcommand{\tabposratiowknn}{\APS(KNN)}
\newcommand{\tabthposratio}{\APS(threshold)}
\newcommand{\signmark}{*}

\subsection{\Binaryclass}
\label{sec:binaryclass}
In our setting, we can also incorporate class relations by  directly
altering the output of a binary classifier, as follows.
We first train a
standard SVM, 
treating ratings greater than 0.5 as positive labels and others
as negative labels.
If we then consider the resulting classifier to output a
{\em positivity-preference function} $\pospref{\doc}$,
we can then learn a series of thresholds to convert this value
into the desired label set, under the assumption that 
the bigger $\pospref{\doc}$ is, the more positive the review.\footnote{
This is not necessarily
true: if the classifier's goal is to optimize binary classification
error, its major concern is to increase confidence in the
positive/negative distinction, which may not correspond to
higher confidence in separating ``five stars'' from 
``four stars''.}
This algorithm always outperforms the majority-class baseline, 
but not to the degree that the best of SVM OVA and SVM regression does.
\newcite{Koppel+Schler:05a} independently found in a three-class study
that thresholding a positive/negative
classifier trained only on clearly positive or clearly
negative examples did not yield large improvements.

\subsection{Discretizing regression}
In our experiments with \svmreg, we discretized regression
output via a set of fixed decision thresholds $\{0.5, 1.5, 2.5, ...\}$
to map it into our set of class labels.
Alternatively, we can 
learn the thresholds instead.
Neither option clearly outperforms the other in the four-class case.
In the three-class setting, the learned  version 
provides noticeably better performance in two of the four datasets.
But these results taken together still mean that in many cases, the difference is negligible,
and 
if we had started down this path, we would have needed to consider similar
tweaks for \svmova as well. 
We therefore stuck with the simpler version in order to maintain focus on the
central issues at hand.


\begin{thebibliography}{}

\bibitem[\protect\citename{Atkeson, Moore, and
  Schaal}1997]{Atkeson+Moore+Schaal:97a}
Atkeson, Christopher~G., Andrew~W. Moore, and Stefan Schaal.
\newblock 1997.
\newblock Locally weighted learning.
\newblock {\em Artificial Intelligence Review}, 11(1):11--73.

\bibsnip\bibitem[\protect\citename{Boykov, Veksler, and
  Zabih}1999]{Boykov+Veksler+Zabih:99a}
Boykov, Yuri, Olga Veksler, and Ramin Zabih.
\newblock 1999.
\newblock Fast approximate energy minimization via graph cuts.
\newblock In {\em Proceedings of the International Conference on Computer
  Vision (ICCV)}, pages 377--384.
\newblock Journal version in {\em IEEE Transactions on Pattern Analysis and
  Machine Intelligence (PAMI)} 23(11):1222--1239, 2001.

\bibsnip\bibitem[\protect\citename{Collins-Thompson and
  Callan}2004]{Collins-Thompson+Callan:04a}
Collins-Thompson, Kevyn and Jamie Callan.
\newblock 2004.
\newblock A language modeling approach to predicting reading difficulty.
\newblock In {\em HLT-NAACL: Proceedings of the Main Conference}, pages
  193--200.

\bibsnip\bibitem[\protect\citename{Das and Chen}2001]{Das+Chen:01a}
Das, Sanjiv and Mike Chen.
\newblock 2001.
\newblock Yahoo! for {Amazon}: Extracting market sentiment from stock message
  boards.
\newblock In {\em Proceedings of the Asia Pacific Finance Association Annual
  Conference (APFA)}.

\bibsnip\bibitem[\protect\citename{Dave, Lawrence, and
  Pennock}2003]{Dave+Lawrence+Pennock:03a}
Dave, Kushal, Steve Lawrence, and David~M. Pennock.
\newblock 2003.
\newblock Mining the peanut gallery: Opinion extraction and semantic
  classification of product reviews.
\newblock In {\em Proceedings of WWW}, pages 519--528.

\bibsnip\bibitem[\protect\citename{Engstr\"om}2004]{Engstroem:04a}
Engstr\"om, Charlotta.
\newblock 2004.
\newblock Topic dependence in sentiment classification.
\newblock Master's thesis, University of Cambridge.

\bibsnip\bibitem[\protect\citename{Herbrich, Graepel, and
  Obermayer}2000]{Herbrich+Graepel+Obermayer:00a}
Herbrich, Ralf, Thore Graepel, and Klaus Obermayer.
\newblock 2000.
\newblock Large margin rank boundaries for ordinal regression.
\newblock In Alexander~J. Smola, Peter~L. Bartlett, Bernhard {Sch\"olkopf}, and
  Dale Schuurmans, editors, {\em Advances in Large Margin Classifiers}, Neural
  Information Processing Systems. MIT Press, pages 115--132.

\bibsnip\bibitem[\protect\citename{Horvitz, Jacobs, and
  Hovel}1999]{Horvitz+Jacobs+Hovel:99a}
Horvitz, Eric, Andy Jacobs, and David Hovel.
\newblock 1999.
\newblock Attention-sensitive alerting.
\newblock In {\em Proceedings of the Conference on Uncertainty and Artificial
  Intelligence}, pages 305--313.

\bibsnip\bibitem[\protect\citename{Ishikawa}2003]{Ishikawa:03a}
Ishikawa, Hiroshi.
\newblock 2003.
\newblock Exact optimization for {Markov} random fields with convex priors.
\newblock {\em IEEE Transactions on Pattern Analysis and Machine Intelligence},
  25(10).

\bibsnip\bibitem[\protect\citename{Ishikawa and Geiger}1998]{Ishikawa+Geiger:98a}
Ishikawa, Hiroshi and Davi Geiger.
\newblock 1998.
\newblock Occlusions, discontinuities, and epipolar lines in stereo.
\newblock In {\em Proceedings of the 5th European Conference on Computer Vision
  (ECCV)}, volume~I, pages 232--248, London, UK. Springer-Verlag.

\bibsnip\bibitem[\protect\citename{Joachims}1999]{Joachims:99a}
Joachims, Thorsten.
\newblock 1999.
\newblock Making large-scale {SVM} learning practical.
\newblock In Bernhard Sch\"{o}lkopf and Alexander Smola, editors, {\em Advances
  in Kernel Methods - Support Vector Learning}. MIT Press, pages 44--56.

\bibsnip\bibitem[\protect\citename{Kleinberg and Tardos}2002]{Kleinberg+Tardos:02a}
Kleinberg, Jon and \'Eva Tardos.
\newblock 2002.
\newblock Approximation algorithms for classification problems with pairwise
  relationships: {Metric} labeling and {Markov} random fields.
\newblock {\em Journal of the ACM}, 49(5):616--639.

\bibsnip\bibitem[\protect\citename{Koppel and Schler}2005]{Koppel+Schler:05a}
Koppel, Moshe and Jonathan Schler.
\newblock 2005.
\newblock The importance of neutral examples for learning sentiment.
\newblock In {\em Workshop on the Analysis of Informal and Formal Information
  Exchange during Negotiations (FINEXIN)}.

\bibsnip\bibitem[\protect\citename{Liu, Lieberman, and
  Selker}2003]{Liu+Lieberman+Selker:03a}
Liu, Hugo, Henry Lieberman, and Ted Selker.
\newblock 2003.
\newblock A model of textual affect sensing using real-world knowledge.
\newblock In {\em Proceedings of Intelligent User Interfaces (IUI)}, pages
  125--132.

\bibsnip\bibitem[\protect\citename{McCallum}1999]{McCallum:99a}
McCallum, Andrew.
\newblock 1999.
\newblock Multi-label text classification with a mixture model trained by {EM}.
\newblock In {\em AAAI Workshop on Text Learning}.

\bibsnip\bibitem[\protect\citename{McCullagh}1980]{McCullagh:80a}
McCullagh, Peter.
\newblock 1980.
\newblock Regression models for ordinal data.
\newblock {\em Journal of the Royal Statistical Society}, 42(2):109--42.

\bibsnip\bibitem[\protect\citename{Pang and Lee}2004]{Pang+Lee:04a}
Pang, Bo and Lillian Lee.
\newblock 2004.
\newblock A sentimental education: Sentiment analysis using subjectivity
  summarization based on minimum cuts.
\newblock In {\em Proceedings of the ACL}, pages 271--278.

\bibsnip\bibitem[\protect\citename{Pang, Lee, and
  Vaithyanathan}2002]{Pang+Lee+Vaithyanathan:02a}
Pang, Bo, Lillian Lee, and Shivakumar Vaithyanathan.
\newblock 2002.
\newblock Thumbs up? {Sentiment} classification using machine learning
  techniques.
\newblock In {\em Proceedings of EMNLP}, pages 79--86.

\bibsnip\bibitem[\protect\citename{Qu, Shanahan, and Wiebe}2004]{Qu+Shanahan+Wiebe:04a}
Qu, Yan, James Shanahan, and Janyce Wiebe, editors.
\newblock 2004.
\newblock {\em Proceedings of the {AAAI} Spring Symposium on Exploring Attitude
  and Affect in Text: Theories and Applications}.
\newblock AAAI Press.
\newblock AAAI technical report SS-04-07.

\bibsnip\bibitem[\protect\citename{Rifkin and Klautau}2004]{Rifkin+Klautau:04a}
Rifkin, Ryan~M. and Aldebaro Klautau.
\newblock 2004.
\newblock In defense of one-vs-all classification.
\newblock {\em Journal of Machine Learning Research}, 5:101--141.

\bibsnip\bibitem[\protect\citename{Schapire and Singer}2000]{Schapire+Singer:00a}
Schapire, Robert~E. and Yoram Singer.
\newblock 2000.
\newblock {BoosTexter}: A boosting-based system for text categorization.
\newblock {\em Machine Learning}, 39(2/3):135--168.

\bibsnip\bibitem[\protect\citename{Smola and Sch\"{o}lkopf}1998]{Smola+Schoelkopf:98a}
Smola, Alex~J. and Bernhard Sch\"{o}lkopf.
\newblock 1998.
\newblock A tutorial on support vector regression.
\newblock Technical Report NeuroCOLT NC-TR-98-030, Royal Holloway College,
  University of London.

\bibsnip\bibitem[\protect\citename{Subasic and Huettner}2001]{Subasic+Huettner:01a}
Subasic, Pero and Alison Huettner.
\newblock 2001.
\newblock Affect analysis of text using fuzzy semantic typing.
\newblock {\em {IEEE} Transactions on Fuzzy Systems}, 9(4):483--496.

\bibsnip\bibitem[\protect\citename{Takamura, Matsumoto, and
  Yamada}2004]{Takamura+Matsumoto+Yamada:04a}
Takamura, Hiroya, Yuji Matsumoto, and Hiroyasu Yamada.
\newblock 2004.
\newblock Modeling category structures with a kernel function.
\newblock In {\em Proceedings of CoNLL}, pages 57--64.

\bibsnip\bibitem[\protect\citename{Tong}2001]{Tong:01a}
Tong, Richard~M.
\newblock 2001.
\newblock An operational system for detecting and tracking opinions in on-line
  discussion.
\newblock {SIGIR} Workshop on Operational Text Classification.

\bibsnip\bibitem[\protect\citename{Turney}2002]{Turney:02a}
Turney, Peter.
\newblock 2002.
\newblock Thumbs up or thumbs down? {Semantic} orientation applied to
  unsupervised classification of reviews.
\newblock In {\em Proceedings of the ACL}, pages 417--424.

\bibsnip\bibitem[\protect\citename{Vapnik}1995]{Vapnik:95a}
Vapnik, Vladimir.
\newblock 1995.
\newblock {\em The Nature of Statistical Learning Theory}.
\newblock Springer.

\bibsnip\bibitem[\protect\citename{Wilson, Wiebe, and Hwa}2004]{Wilson+Wiebe+Hwa:04a}
Wilson, Theresa, Janyce Wiebe, and Rebecca Hwa.
\newblock 2004.
\newblock Just how mad are you? {Finding} strong and weak opinion clauses.
\newblock In {\em Proceedings of AAAI}, pages 761--769.

\bibsnip\bibitem[\protect\citename{Yu and
  Hatzivassiloglou}2003]{Yu+Hatzivassiloglou:03a}
Yu, Hong and Vasileios Hatzivassiloglou.
\newblock 2003.
\newblock Towards answering opinion questions: Separating facts from opinions
  and identifying the polarity of opinion sentences.
\newblock In {\em Proceedings of EMNLP}.

\bibsnip\bibitem[\protect\citename{Zhu}2005]{Zhu:05a}
Zhu, Xiaojin~(Jerry).
\newblock 2005.
\newblock {\em Semi-Supervised Learning with Graphs}.
\newblock {Ph.D.} thesis, Carnegie Mellon University.

\end{thebibliography}
\end{document}